# A SIMPLE FLOOD FORECASTING SCHEME USING WIRELESS SENSOR NETWORKS

Victor Seal[1], Arnab Raha[1], Shovan Maity[1], Souvik Kr Mitra[1], Amitava Mukherjee[2] and Mrinal Kanti Naskar[1]

[1]Advanced Digital and Embedded Systems Laboratory, Department of Electronics and Telecommunication Engineering, Jadavpur University, Kolkata, India
`victor.seal@yahoo.co.in, arnabraha1989@gmail.com, shovanju35@gmail.com, souvikmitra.ju@gmail.com, mrinalnaskar@yahoo.co.in`
[2] IBM India Private limited, Kolkata, India
`amitava.mukherjee@in.ibm.com`

## ABSTRACT

*This paper presents a forecasting model designed using WSNs( Wireless Sensor Networks) to predict flood in rivers using simple and fast calculations to provide real-time results and save the lives of people who may be affected by the flood. Our prediction model uses multiple variable robust linear regression which is easy to understand and simple and cost effective in implementation, is speed efficient, but has low resource utilization and yet provides real time predictions with reliable accuracy, thus having features which are desirable in any real world algorithm. Our prediction model is independent of the number of parameters, i.e. any number of parameters may be added or removed based on the on-site requirements. When the water level rises, we represent it using a polynomial whose nature is used to determine if the water level may exceed the flood line in the near future. We compare our work with a contemporary algorithm to demonstrate our improvements over it. Then we present our simulation results for the predicted water level compared to the actual water level.*

## 1. Introduction

Floods are responsible for the loss of precious lives and destruction of large amounts of property every year, especially in the poor and developing countries, where people are at the mercy of natural elements. A lot of effort has been put in developing systems which help to minimize the damage through early disaster predictions. As a network for the prediction model has to be deployed in the rural areas, there is a severe limitation of resources like money, power and skilled man-power.

The organization of our work below is as follows. In the remaining of this section, we introduce the different types of forecasting models and WSN deployment models prevalent. In Section 2, we present contemporary work related to our present work. In Section 3, we propose the system architecture and details for our deployment scheme. In Section 4, we present our algorithm and elaborate on all its features and then compare it with a recent work to demonstrate the new features we have incorporated. In Section 5, we provide our Simulation Results to confirm the efficiency of our algorithm. In Section 6, we discuss on possible future work and conclude. Finally we provide our references.

Continuing with the prevalent models, for a long time, two early types of flood detection and warning systems have been established: A) non-WSN and B) WSN

## 1.1. Non-WSN systems:

These systems have a primitive technique of flood detection requiring trained personnel. It involves mostly manual procedures and hence is expensive. Here, the reliability of the entire mechanism depends upon the skill and experience of the personnel employed and is subsequently limited by their speed and agility. Recent developments have led to automated telemetry systems. Even these are expensive as they require periodic installation of repeaters and transmitters. Most of the telemetric systems follow a centralized computational technique. Also deployment of a large number of telemetric systems to cover the entire region becomes impractical due to the large expenses involved. In spite of having these issues, the non-WSN systems are operating in many countries in the world.

## 1.2. WSN systems:

WSNs can be defined as low power, low cost, multi-hopping systems that are independent of external service providers, can form an extendable network without line of sight coverage; but have self-healing data paths. WSNs can be deployed more or less homogeneously in a geographical region using a two-tiered approach having clusters of short distance communicating nodes together with some nodes capable of communication over a wider range. WSN nodes communicate only with neighbouring nodes to reduce the transmission power and losses, thus eliminating the need for expensive repeaters and transmitters used in traditional telemetry systems. Every node in a WSN can act as a data acquisition device, a data router and a data aggregator. This architecture maximizes the redundancy and consequently the reliability of the entire flash- flood monitoring system. The independence from third party providers and the absence of infrastructure requirements – as those needed in cellular based telemetry systems allow a WSN to be deployed quickly. They allow online, self-calibration of the prediction model.

Three types of models may be designed: Centralized, Distributed or Hybrid

## 1.3. Centralized Model

A centralized model is one where computation occurs at the central node only. It needs less number of components as terminal nodes don't need the electronic components required for computational purposes. However, the whole system fails if the central node fails even if it is only a single point of failure (assuming a fully connected backup of the central node is absent). There is also a need to avoid bottlenecks due to transmission of measured data simultaneously from all the sensing nodes, which can limit the data available to the model to use for calibration and prediction thus limiting the flexibility of the model.

Previously, most of the available WSN structured models used a prediction model with a centralized computational technique where we come across three main shortcomings:

i) They require large amount of data for calibration (i.e. they need a huge amount of training and verification data) for prediction of the unknown statistical coefficients.

ii) The computational algorithm is complex which in turn increases the computational power as well as time.

iii) System reliability is reduced as there are no redundancy checks to provide corrections in case a value at the central node becomes corrupted during calculations.

## 1.4. Distributed Model

A distributed model is one with computations at several levels instead of only one computing node as in the previous model.

Most of the disadvantages of a centralized system are addressed in this model as different terminal sensors act together to provide the collected data to bridge nodes, identify internal failures and to adapt to changes in topology. This model also increases the reliability of a system by introducing redundancy as the same calculations are done at different nodes and then matched. However, they suffer from the difficulty and cost of maintenance of a number of sensors with additional functionalities and also in ensuring proper communication among these heterogeneous sensors.

## 1.5. Hybrid Model

As the name suggests, this model comprises part of the centralized model and part of the distributed model. The extent to which a part of each model (centralized or distributive) is included in it is flexible and may vary as per system requirements. This model aims to partly combine the advantages of both the above systems and simultaneously, cancel out the disadvantages of either system by combining both. Though the network hierarchy remains the same here as in the previous cases, computation is done on multiple nodes unlike the centralized method, and yet computation in almost all nodes is not done as in the distributed method. Thus, it is possible to bring about an appropriate balance of the cost of deployment and the redundancy and reliability of the system as needed.

# 2. Related Work

Nowadays, almost all the flood forecasting systems available are WSN-based as in [1, 3, and 5]. Centralised Non-WSN based telemetric systems fail to ensure fast or reliable warning and thus have few or no forecasting methods employed using them. In majority of the data-driven statistical flood-prediction models, details of the landscape, soil composition, and land cover, along with atmospheric conditions and hydro-meteorological measurements like soil moisture are required as in [1, 3]. As seen in [3, 5], the development of both lumped and spatially distributed models are common for the analysis of such rainfall-runoff measuring systems. Several algorithms like those in [1, 4] which attempt to serve the same purpose as ours requires an exhaustive set of tests for calibration and require repeated re-calibrations, making their process power consuming and hard to achieve in real time. Some algorithms like [4] make their models rigid by restricting the number of parameters they consider for flood forecasting to a fixed quantity.

Reference [7] suggests a distributed node deployment to ensure proper communication in a WSN quite similar to ours. References [4] and [6] handle issues in parameter data management and system design and comparison of various distributed models. References [2, 10-13] elaborate on robust linear regression –the mathematical base for calculations our algorithm and discuss in details its advantages comparing it to simple linear regression in [2]. Reference [2] also provides standard functions for assigning weights while using the weighted regression model suggested in our paper. Reference [13] elaborates on outlier detection as an extensive problem of robust regression. Reference [9] discusses forecasting in general. It provides an exhaustive analysis of the various types of forecasting models, simple and complicated.

# 3. Proposed Work

## 3.1. System Architecture for our WSN:

The proposed WSN system architecture for flood forecasting consists of sensors(which sense and collect the data relevant for calculations), some nodes referred to as computational nodes (that have large processing powers and implement our proposed distributed prediction algorithm) and a manned central monitoring office(which verifies the results with the available online information, implements a centralized version of the prediction algorithm as a redundancy mechanism, issues alerts and initiates evacuation procedures). Different types of sensors are required to sense water discharge from dam, rainfall, humidity, temperature, etc. The data collected by these sensors are used in the flood prediction algorithm. The computational nodes possess powerful CPUs required to implement the distributed prediction model. The computational nodes are supposed to communicate the prediction results to the monitoring node. They also have communication between themselves for detecting malfunctioning of nodes. The work of the central node is not our concern. However it is important to include a manned central node in this whole process to raise the alarm and co-ordinate evacuation measures if needed.

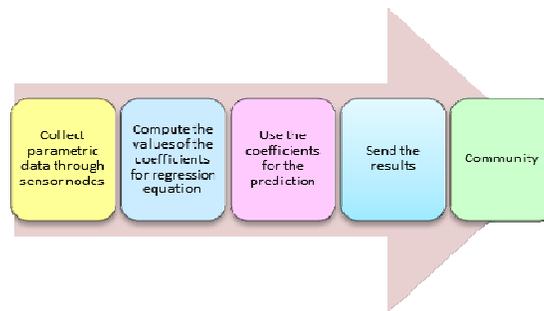

Figure 1: Flow chart for the transfer of data and information from sensors to the human community

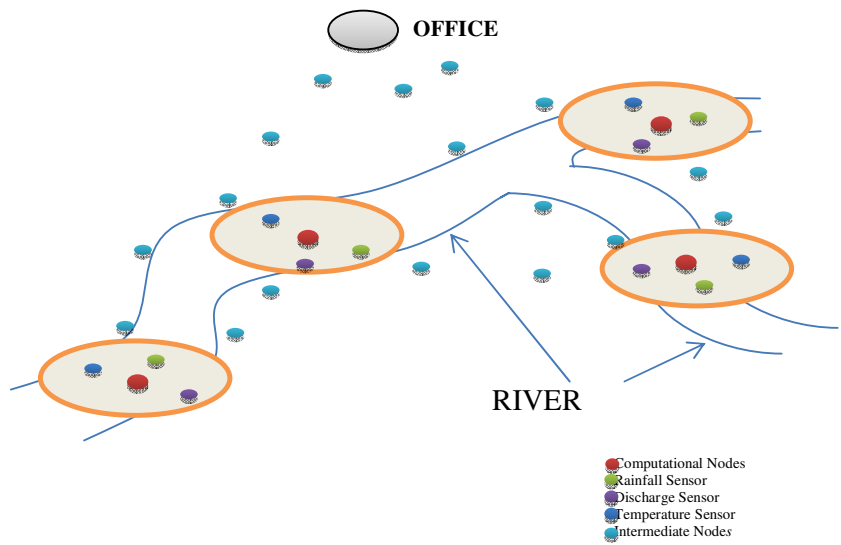

Figure 2: WSN Deployment Scheme in a flood-prone river basin.

The next important aspect is to minimize the effect of a node failure while connecting the computational nodes to the central. Intermediate nodes (INs) have to be deployed to ensure this connectivity in case the central does not fall within the communication range of all the nodes. Given below is the entire picture of the WSN at site is given. As we can see the river has been broken into several monitoring zones. In each zone, a sensor node collects data and sends them to its computational node. Data collection and localized prediction takes place at each computational node. The computational nodes then send the data to the central (office) node and also share it among themselves.

## 3.2. Procedure

We have considered our prediction algorithm to be compatible with a 2 layered distributed framework based network architecture which consists of a number of sensor nodes collecting different data required for prediction through temperature, rainfall sensors and others. These data is communicated to the single computation node (or an administrative office as the case may be) which acts as the cluster head bearing the responsibility of most of the computations and predictions (both online and offline) required. Obviously, it is supplied with greater computational power and processing capability. With data collected online dynamically, forecasts are made using our suggested algorithm and the results are communicated to the local administrative office which raises the flood alarm.

The instants at which sensors read data are decided by any one or more of these four parameters:

### 3.2.1. Time

This is the most commonly used parameter, giving an easily understood technique. The system supervisor decides a time interval after which a reading is to be taken, accordingly a sensor takes the time driven data as input. In our algorithm we use a time set instead of a specific interval, as discussed under the time multiplier function.

### 3.2.2. Event

Event refers to sudden change in a parameter affecting the desired output. Suppose the next time driven data is slated to be taken after an hour. But a dam has started discharging water at such a rate that flood may occur within the next 30 minutes. A hardware interrupt has to override the time driven mechanism to get an instant event driven data and make a reliable prediction before the flood occurs.

### 3.2.3. Query

This parameter comes to effect if an administrator wants to see the immediate level and get an instantaneous prediction even when it's not time to take a reading. A hardware interrupt is used get the instant query driven data. It is quite similar to the event driven case, the only difference being that this interrupt is provided by a human unlike natural parameters that trigger the event triggering interrupt.

### 3.2.4. System Interrupt

Unlike event or query driven interrupts which are both external events, system interrupt is internal, generated by a computational node or the office node itself when there is some corruption or loss of data during any phase of the communication process-data sensing or recording, transmission, accumulation or calculation. The office node or computational nodes send a request to the sensor nodes to take another readings (or set of readings) whenever it detects an error by using a system interrupt.

The algorithm suggested in our paper takes care of these four parameters. It operates using a time driven mechanism and checks for hardware interrupt at every instant of wait interval (time interval before next reading) and forfeits its wait state if there is an interrupt. Our algorithm is supposed to run on each of the computational nodes and also in the office node. The latter is optional and may be used as a provision for redundant checking to reduce the possibility of false prediction and keep the alert generation to a minimum. Rainfall, discharge, temperature etc. data are collected from the sensor nodes and communicated to the computational node. These computational nodes transmit the information collected from the sensor nodes and also their own predictions to the office node with the help of the intermediate nodes. The office reruns the algorithm on its own and generates the alert if needed.

It is necessary for our algorithm to be effective that the computational nodes are provided with previous data needed to standardize the coefficients needed for fitting the equation. Although such data is required for initial calibration only, subsequent updating and recalibration of the coefficients are done through the live online data received from the sensor nodes deployed at the site. The test data required to calibrate the coefficients of the least square fit are to be stored in a database embedded in the memory of the sensor nodes (computational nodes). This database gets updated regularly at intervals specified earlier according to the prevalent environmental conditions. Greater sampling rate may be used during severe environmental conditions like thunderstorms, cyclones and hurricanes when chances of flood are high.

The benefit of such a technique is that it can integrate a large number of parameters upon which the river water-level varies providing us with a more precise estimate and minimising the possibility of false alarms. For the sake of simplicity, we, at the beginning, choose a threshold level for river water such that if the predicted value is more than that level an alarm can be raised. Choosing such a threshold level is very critical and imperative in our case as it may lead to unnecessary false alarms if the threshold level is inaccurate. It is certainly necessary that correct information regarding such threshold level should be given by the river maintenance and administrative office.

If the river basin characteristics show that floods have occurred at a level of say 30m then we can set a threshold value $L_{TH} = 25$, such that if the prediction shows a value higher than $L_{TH}$ we may suggest a potentially dangerous situation. Now suppose the parameters vary such that water level remains around that 25m mark for 10 days in a row. The algorithm would be inefficient it gives a false time every time the water level crosses the threshold. Instead we suggest reducing the sampling interval in such a situation so that the system sensitivity increases and it sounds an alarm only if needed. Even the intensity of flood may be determined based on the predicted value.

To make the prediction procedure faster and more accurate, we have to decide the importance of a parameter in predicting the output. For example, water level may depend on rainfall, discharge from dam, etc. but number of ferries across the river is surely not a parameter for predicting its water level.

# 4. Our Prediction Model:

We present our generalised algorithm here which runs in the node(s) where the computation is to be done for flood prediction.

## 4.1. Program Algorithm:

We provide the algorithm of the main function first and follow it up with the part wise algorithms of the different functions used.

**Input:** Previous data tables: L=water level, R= rainfall, D=Discharge, $X_i$ ,i=1,2…= other parameters affecting water level; SC=Storage capacity (in number of readings) of node. S=stored regression coefficient array, W=weight assigned to S. P=present coefficient matrix.

Initialise: Set counter=1, weight=0, initialise all tables to zeroes.

**Output:** Expected value of L.

**Step 1**. Read past data tables: L=water level, R= rainfall, D=Discharge, $X_i$, i=1,2…=other parameters affecting water level.

**Step 2.** Compute/re-compute linear regression model (i.e. coefficient matrix P) relating L with the above parameters using the robust fit function.

**Step 3.** Find the weighted mean of P and S (weight of P=counter value now, weight of S=W).This is our final regression model.

**Step 4.** Input the present value of parameters and predict corresponding value of L based on the final regression model.

**Step 5.** Decide time interval before next reading using the time multiplier function.

**Step 6.** Maintain a table of water levels v/s time and see if the final trend(most recent reading to back in time) is increasing or not.

**Step 7.** If water level trend is not increasing, flood is not expected. Go to step 10.

If increasing, use Quadratic Fit function to approximate the trend v/s time (from present value to the 1st local minima reached going back in time) and generate a prediction model for it.

**Step 8.** Put in values of time in future in progressive order into the prediction model above and predict future values of water level.

**Step 9.** If water level stays below flood line up to a defined reliability period of our prediction model, predict flood not possible.

If water level exceeds the flood line within the defined reliability period, predict flood possibility.

**Step 10.** Recalibrate time interval of step 3 based on this prediction.

**Step 11.** Input present value of water level. Increment counter by 1.

**Step 12:**If counter value is less than SC, append all present values to the end of their respective tables in step1(L, D, R, $X_i$) . Then go to step 2.

Else, store the coefficient matrix and increment its weight (W) by SC. Then clear all the data and output tables, reset counter to one to ready it for taking inputs.

**Step 13:** Enter a wait state; waiting time= recalibrated time interval (step 10)

**Step 14:** Check-is the system still in wait state?

If yes, go to step 15.

Else go to step 2.

**Step 15:** Check- is there any hardware interrupt (event/query driven data acquisition demand) now?

If yes, go to step 2 immediately.

Else go to step 14.

## 4.2. Quadratic Fit function:

We approximate the increasing water level by a second order polynomial function to predict future values i.e. $y = a_1x^2 + a_2x + a_3$. Where, x is a time index starting from the moment water level starts rising;

y=corresponding water level at that time.

Tabulating for several values $(x_i, y_i)$, the solution is given in form:

$Y_{nx1} = X_{nx3}A_{3x1}$ (n=no. of readings) where X is the Vandermonde matrix whose elements are given by

$X_{i,j} = x_i^{3-j}$ (i=1,2,3;j=1,2,3) i.e. simply stated, the elements of each row I are $x_i^2, x_i, 1$ respectively.

In Y=XA we know Y, X; solving for A gives the coefficient matrix $[a_1\ a_2\ a_3\ ]'$ above. To solve,

$X^TY = X^TXA$

Or, $(X^TX)^{-1}\ X^TY = A$.

Now we have the $2^{nd}$ order polynomial equation $y = a_1x^2 + a_2x + a_3$ ready to use.

Going by the present nature of the curve, we predict its future trend and see if it goes over the flood line within the reliability period of our future prediction and predict accordingly.

The $2^{nd}$ order polynomial function ensures simplicity of calculation over larger polynomial.

## 4.3. Robust Fit function:

Using ordinary least-squares fitting has the main problem in its sensitivity to outliers. Squaring the residuals magnifies the effects of these extreme data points and hence they have a large influence on the final curve. To minimize the influence of outliers, we fit our data using robust least-squares regression.

Robust fitting with poly-square weights (explained later) uses an iteratively reweighted least-squares algorithm, and follows this procedure:

1. Initially, we approximate the model by weighted least squares using poly-square weight minimization:

2. Secondly, we compute the adjusted residuals and give them a typical value. The adjusted residuals are given by

$r_{adjusted} = r_i / (sqrt(i-h_i)$

$r_i$ are the usual least-squares residuals and $h_i$ are weights that adjust the residuals by down-weighting high-leverage data points (more specifically, outliers), which have a large effect on the least-squares fit. The standardized adjusted residuals are given by

$$u = r_{adjusted} / K$$

$K$ is a tuning constant equal to 4.685, and $s$ is the robust variance given by $MAD/0.6745$ where $MAD$ is the median absolute deviation of the residuals. Mathematically,

$MAD = median_i (|X_i - median_j(X_j)|)$

1.  We calculate the robust weights as a function of $u$. The poly-square weights are given by

$w_i =$    $(1-(u_i)^2)^2$        if $|u_i| < 1$

          $0$                if $|u_i| >= 1$

2.If the fit converges within a defined tolerable range, our work is done. Otherwise, we perform the next iteration of the fitting procedure by returning to the first step.

## 4.4. Poly-square weight minimization:

This method is used to minimize a weighted sum of two or more squares of numbers. A bi-square number is the weighted sum of the squares of two numbers. Extending this definition, a poly-squared number may be defined as the weighted sum of squares of multiple numbers. The weight given to each data point (each number in the sum) depends on how far it is from the fitted line.

$y = a_1 + a_2x_1 + a_3x_2 + \ldots + a_{m+1}x_m$ is a general form of m variable regression. For n sets of values {y(i) ,$x_1(i)$ ,$x_2(i)$ ,…,$x_m(i)$ }, i=1,2,3…n; we have n equations:

$y(i) = a_1 + a_2x_1(i) + a_3x_2(i) + \ldots a_{n-1}x_n(i)$

In matrix form, it is $Y_{nx1} = X_{nxm}A_{mx1}$ where X matrix has 1st column of1s,2nd column of $x_1$ values (i.e. 1st row of 1st of n readings,…ith row=ith reading of $x_1$),3rd column of $x_2$ ,ith column of $x_{i-1}$ and the last column (m+1th column) of $x_m$ values.

The central point of poly-square weight minimization is, unlike simple least square minimization, we do not give same value/weight to the equation $y = a_1 + a_2x_1 + a_3x_2 + \ldots + a_{m+1}x_m$ for every value of i.Let $m_y$ denote mean of $y_i$ .The weight assigned to the ith equation is determined by the value of $(y_i - m_y)^2$. We have used the Andrews function for our procedure but any of the weight assigning functions as suggested in figure2 of [2] may be tried out. For region specific application, the user may try out different functions to decide upon the best predictor for a particular set of data.

In matrix form,

WY=WXA. Where the weight matrix must be nxn, with all the elements  $w_{i,i}$=weight assigned to its corresponding $y_i$ ;all elements except lead diagonal elements=0.

We have to solve for the coefficient matrix A; We know W,Y and X. So, Solving for A:

$X^TWY=X^TWXA$

Or, $(X^TWX)^{-1} X^TWY=A.$

The matrix A found here is used for the next robust approximation.

This method is alterable based on available resources and memory space. We can easily limit the number of iterations to enhance system simplicity by just adding a counter to the algorithm and checking its value at the end of every fit. As an approximation system improves, the structure needed to support it becomes more complicated. For high end systems, non-linear regression models may even be used. For simple sensors, just 2 or 3 iterations of the above algorithm is sufficient to provide a reasonably good approximation over simple least square fit considering how simple the system is.

### 4.5. Time Multiplier function

This function decides the time interval between two successive readings. This is a worst case predictor that tells how soon, or how late the next reading must be taken to predict flood successfully based on two parameters:

#### 4.5.1. Difference in water level between subsequent readings

Say the water level has increased by an amount equal to 90% of flood line between two subsequent readings. Our predictor assumes the increase is not uniform over the interval, but may have happened in the last few moments. Thus it asks for another reading in the minimum possible time to see the gradient and make a better prediction. (Note: Flood is not predicted just if the parameter value is high, we only ask for another reading faster for accurate prediction.)

#### 4.5.2 Present water level

Say the water level has reached 90% of flood line. A little rain/discharge any moment can trigger a flood. So it is best to take readings as soon as possible and make accurate predictions. Continuing the thought process, we assign different time intervals for subsequent readings as logical from the measured/calculated values of the two parameters above.

We define a *time set* containing different time intervals at which the sensor nodes can take readings. Based on the above discussion, the time multiplier function decides which value of the time set is to be taken for a particular calculation.

This function decides the time interval based on a worst case scenario but it doesn't affect prediction of flood. A worst case flood predictor would generate numerous false alarms, hence is undesirable. This implies that our function asks for readings faster with decreasing wait times between successive readings when there is any chance of a flood, and only generates an alarm when there is a realistic chance of flood within a certain time limit, as discussed in the following section.

### 4.6. Recalibration of time interval

Assume the next reading is scheduled to be after an hour. But our algorithm forecasts a flood within twenty minutes from now. Clearly, the time interval must be altered so that it is at least less than the time at which flood is predicted. If a flood is predicted before the minimum time interval in the time set, we divide the entire time set by an integral value, so that the values of the new time set are reduced by that integral factor. Let 'T' be the maximum time interval needed to transmit information to all and act when a flood alarm is generated plus an extra safety time (to account for delays, etc.). The time interval in prediction may be reduced up to time T. If a flood is predicted before this time T, an alarm must be generated.

## 4.7. Consideration for node storage capacity

Our algorithm stores the regression coefficient values and frees the system memory by removing parameter values whenever there is a constraint on system memory. Effectively, nothing is lost as the single coefficient matrix is the result desirable from such enormous amounts of parameter data. A new coefficient matrix is constructed using the new inputs. A resultant coefficient matrix calculated using the weighted mean of these two, each matrix being assigned a weight in proportion to the number of readings whose values it incorporates.

## 4.8. Error Measurement

Using the root mean square (rms) value of the errors in all readings as a figure of merit for our algorithm is faulty since it is susceptible to outliers. Suppose in a particular data set, the sensor reading the present water level submits a corrupted value i.e. there is a severe measurement error for that reading. Even if the prediction is close to the actual water level, the system would record a huge error due to the measurement error. Then the rms error found will be quite high due to a single high value. Therefore we suggest using a weighted root mean square (wrms) method for determining the accuracy of the system. A negligible weight is assigned to such outliers in the error values so that their effect is nullified. Tis method would depict the system performance more reasonably. An added advantage of this method is that the algorithm can use it to phase out inconsistent or corrupt data from the system, especially when memory constraints are reached. That is, when some data has to be deleted to incorporate new readings, the weights wrms method assigns to an error is an indication of the reliability of that reading and data may be deleted accordingly. Outlier detection is an extensive problem in itself and discussed in [13].

## 4.9. Power Consumption Optimization for deployed sensors

In our prediction model, a lot of power can be saved if we keep the sensors in sleep mode when they are in a wait interval between readings. The nodes would wake up again at the end of the wait interval or if an interrupt is provided in-between to take another reading. The power saving mode required in our case is such that it turns off the transducer (sensing) part and the transmitting part of the node while the receiving part is kept on to check for query-driven interrupts or a signal from the computational nodes to take the next reading. This method also ensures that individual sensors need not have a timer for storing and adhering to the wait time but they simply take readings at the end of each wait interval when the computational node sends them a signal (or interrupt). This makes the design of sensors simpler allowing more power savings. Some nodes with minimal power consumption and whose sensing modes are always kept ON have to be deployed at regular intervals to keep checking for external i.e. event driven interrupts. Even in this case, more power savings can be made by turning off the receiving mode of these sensors and the transmission mode is in a sleep mode and is activated whenever there is an abrupt change in the sensed value.

## 4.10. Improvements over [1]:

### 4.10.1. Inclusion of time revision factor

Reference [1] does not talk about forecast time interval for next reading/flood prediction. But we have introduced a logical time predictor to decide the next time for taking a reading. Say a severe drought is going on around the river bed presently with almost no water stored in the dam. It is impractical to ask for data every 5 minutes now. Our algorithm handles these issues.

### 4.10.2. Enhancement of the flood dependency on its parameters

Reference [1] proposes to include prediction error values as a parameter along with rainfall, etc. and other parameters on which flood level depends, to recalibrate the regression based model and predict a flood level with better accuracy. Treating error as a parameter affecting water level is a poor technique. Also, the parameters will get considerably worsened in presence of any outliers. We append the present input values to our data tables to do the same recalibration using robust techniques which are quite immune to outliers and still have nearly the same simplicity.

### 4.10.3. Use of a weighted linear model

Reference [1] suggests smoothing the data using a low pass filter. Arguably, a much better and formal technique is to assign weights to each data and compute accordingly. Its compatibility with low end systems has already been discussed.

### 4.10.4. Consideration for node storage capacity

As elaborated previously, our algorithm effectively uses all the past data though the actual data is deleted when it exceeds system memory capacity. Reference [1] does not talk about system memory constraints.

### 4.10.5. Independency w.r.t number of parameters

Ideally, our algorithm can handle any number of parameters and data elements, the only restriction being memory and/or available resources. Reference [1] does not talk about this issue.

### 4.8.6. Consideration for power optimization of deployed network

Reference [1] does not discuss anything towards power management of the sensor nodes which is an important factor in any battery-dependent application. We suggest a novel sleep mode for the sensors when not in use as already discussed.

# 5. Simulation Results

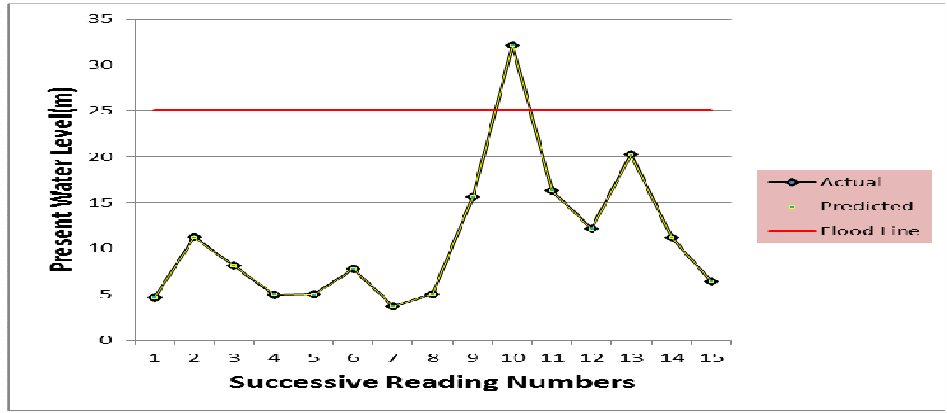

Figure 3. Plot showing the actual and predicted values of present water levels for fifteen successive readings

Our simulation was performed in MATLAB. The predicted values of river-levels are found to be extremely close to those measured from the sensors. Therefore, only one line can be seen for both the actual and the predicted lines. Only once the river water level crossed the flood line signalling the onset of flood as shown in Fig. 3. False alarms are not generated in any predicted values justifying the reliability of our scheme. Due to lack of actual real time data showing variation of several different parameters varying with the water level, we have simulated the regression model for water level with reference to two parameters only-rainfall and discharge from dam.

Now we calculate the difference in the actual and predicted values to find the error. We take the modulus of the error (so that the same error value, 'x' is shown if the predicted value is either more than or less than the real value by 'x'). Then we normalize the error, i.e. find the error as a percentage of the flood line (25 metre for our simulation). The figure below shows the percentage errors in predicted water level at different time instants compared to the actual water level measured at those instants.

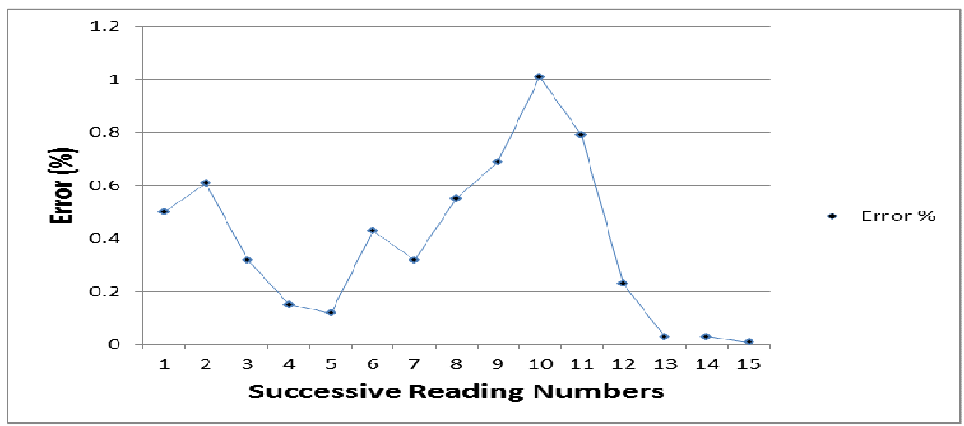

Figure 4. Percentage Error in Predicted Water Level compared to the measured water level for fifteen successive time instants. (The error is normalized with respect to the flood line).

As seen above, the mean error is about 0.4% with the maximum error less than 1.1% which is excellent for any forecasting algorithm. The weighted root mean square error is found to be around 0.36%. Finding the weighted root mean square error also helped us identify readings in which data was corrupted, thus giving minimum or no weightage to those readings.

## 6. Conclusion and Future Work

The forecasting model demonstrates very precise results in forecasting critical parameters for flood when the data collected are accurate. Even if some data entered may be wrong (due to sensor failure at that instant or loss of data in transmission, etc.), the robust fit procedure (or a few iterations of poly-square weighted minimization for very simple nodes) is sufficient to provide a practically accurate result by assigning negligible weights to the outliers. Even when the node storage capacity is reached, our algorithm outlines a novel way to include the results from those data in calculations though the actual data is deleted. The advantages of our algorithm over present models are evident from the comparison with [1] and the simulation results following it. Although real hardware implementation is going to be the next major step in developing this algorithm further, we can also improve upon accurate time prediction by manipulating the prediction algorithm to suit specific rivers.

Future work involves performing field tests to observe the communication process between the nodes and the real-time implementation of the distributed prediction algorithm in situ. Improving the algorithm by power optimization of the sensors for using their battery life more efficiently is another important future work. Using different functions to approximate the trend of increasing water-level, to better approximate the curve, and consequently make better future predictions is yet another avenue for work. Finally, fine-tuning of the algorithm to make it better suited or better adaptable to certain situations, or better still, designing different variants of the algorithm for different applications should also be attempted.

## Authors

Victor Seal is a fourth year undergraduate engineering student, presently pursuing his B.E in Electronics and Telecommunication Engineer ing, Jadavpur University, India. His current research interests include disaster management, wireless sensor networking, ad-hoc networks and communications, embedded system design and control systems.

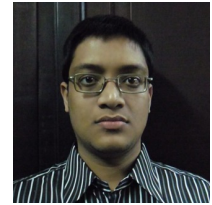

Arnab Raha is a fourth year undergraduate engineering student, currently pursuing his B.E in Electronics and Telecommunication Engineering, Jada vpur University, India. His research interests include wireless sensor networks and ad-hoc networks.

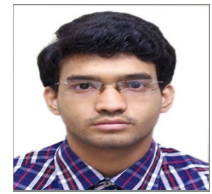

Shovan Maity is a fourth year undergraduate engineering student, currently pursuing his B.E in Electronics and Telecommunication Engineering, Jadavpur U niversity, India. His research interests include disaster management and wireless sensor networks and communications.

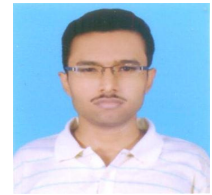

Souvik Kr. Mitra is a fourth year undergraduate engineering student, currently pursuing his B.E in Electronics and Telecommunication Engineering, Jadavp ur University, India. His research interests include wireless sensor networks.

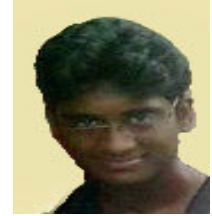

Amitava Mukherjee is a senior manager of IBM GBS India Pvt Ltd from 1995. And has more than 25 years of experience in leading and managing people, program and project in consulting business and collaborative research program. His consulting experiences include Global Delivery, IT strategy consulting and implementation in SAP and Custom Software, software quality management and infrastructure management, resource allocation and hiring, and corporate/int ernal training. Amitava Mukherjee had run and is running scientific research projects funded by government/research council with national and international research experiences cover the areas of wireless communication, mobile computing and communication, sensor network, pervasive computing and mobile governance, optical network, combinatorial optimization, distributed system. He had been on sabbatical from IBM India (Jan 2003-Mar 2005), and visited University of New South Wales, Sydney as visiting Professor (2003-2004) and Royal Institute of Technology, Stockholm as Senior Researcher (2004-2005). From 1983 to 1995, he was with the Department of Electronics and Telecommunication Engineering, Jadavpur University, Calcutta, India. He has over 120 published papers in journals and conference proceedings of international repute (like IEEE, ACM, etc.) and five books in wireless communication and societal engineering. He has the collaboration with different research labs/institutes across continents. Amitava Mukherjee is Ph. D in Computer Science and Engineering, Jadavpur University.

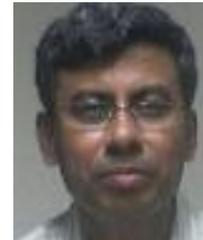

Mrinal Kanti Naskar received his B.Tech. (Hons) and M.Tech degrees from E&ECE Department, IIT Kharagpur, India in 1987 and 1989 respectively and Ph.D. from Jadavpur University, India in 2006.. He served as a faculty member in NIT, Jamshedpur and NIT, Durgapur during 1991-1996 and 1996-1999 respectively. Currently, he is a professor in the Department of Electronics and Tele-Communication Engineering, Jadavpur University, Kolkata, India where he is in c harge of the Advanced Digital and Embedded Systems Lab. His research interests include ad-hoc networks, optical networks, wireless sensor networks, wireless and mobile networks and embedded systems. He is an author/co-author of the several published/accepted articles in WDM optical networking field that include "Adaptive Dynamic Wavelength Routing for WDM Optical Networks" [WOCN,2006], "A Heuristic Solution to SADM minimization for Static Traffic Grooming in WDM uni -directional Ring Networks" [Photonic Network Communication, 2006],"Genetic Evolutionary Approach for Static Traffic Grooming to SONET over WDM Optical Networks" [Computer Communication, Elsevier, 2007], and "Genetic Evolutionary Algorithm for Optimal Allocation of Wavelength Converters in WDM Optical Networks" [Photonic Network Communications,2008].

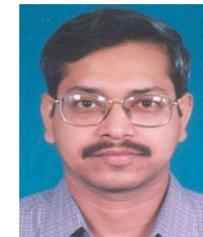